\documentclass[10pt,twocolumn,letterpaper]{article}

\usepackage{cvpr}              % To produce the CAMERA-READY version
\usepackage{cuted}
\usepackage{caption}
\usepackage{float}

\usepackage{array}
\definecolor{cvprblue}{rgb}{0.21,0.49,0.74}
\usepackage[pagebackref,breaklinks,colorlinks,allcolors=cvprblue,hypertexnames=false]{hyperref}

\def\confName{CVPR}
\def\confYear{2027}

\title{DexLatentWorld: 3D Cross-Embodiment Latent Action World Models for Zero-Shot Robotic Manipulation}

\author{
\textbf{Vinay Udit Mohan,\quad Xuchuan Huang,\quad Rolandos Alexandros Potamias}\\[2pt]
Embodied Dexterity Lab (DexLab)\\
Imperial College London
}
\begin{document}
\maketitle

\begin{strip}
    \centering
    \includegraphics[width=0.9\textwidth]{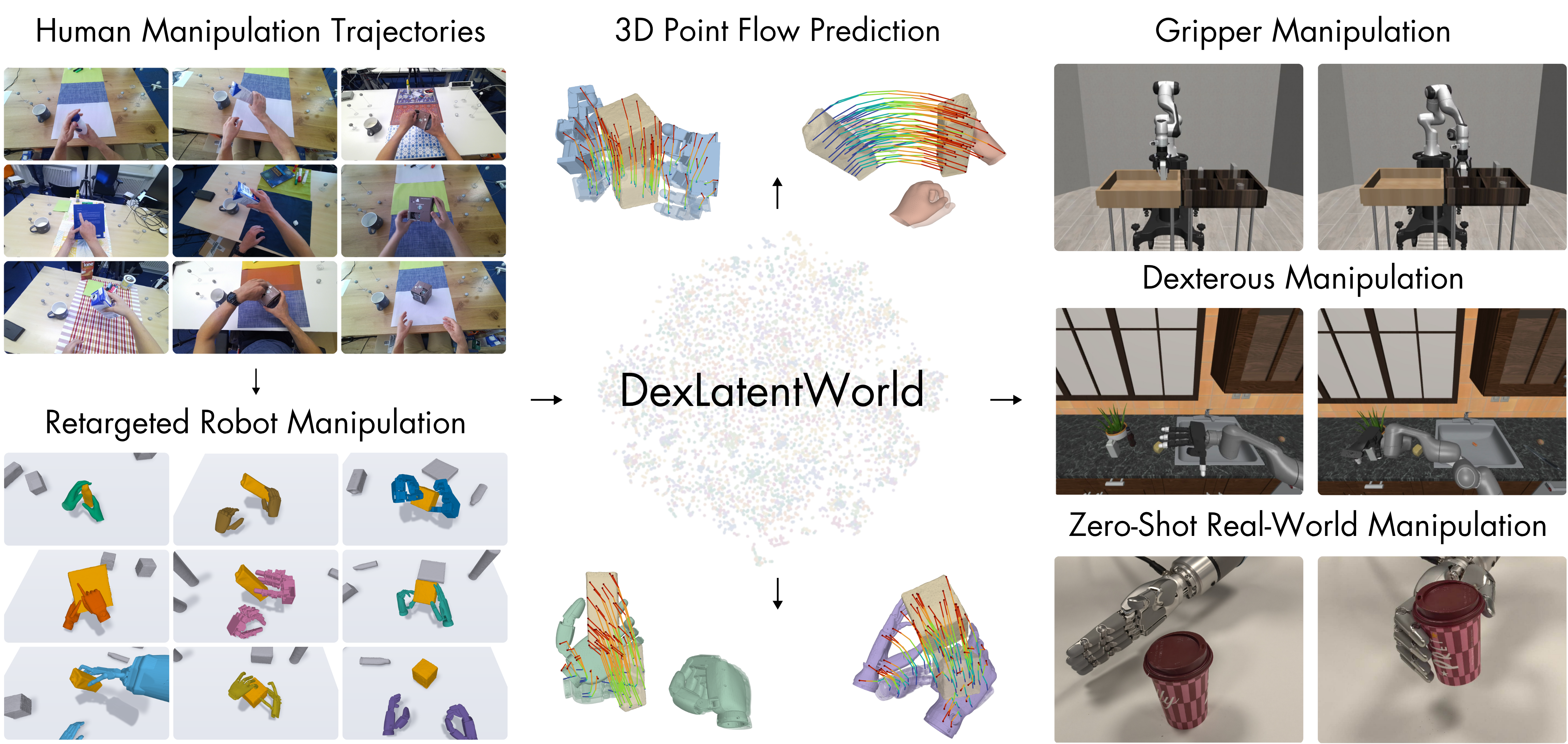}
    \captionof{figure}{We introduce a jointly trained 3D latent-action world model that predicts full-scene point evolution conditioned on cross-embodiment latent actions learned from action-unlabelled manipulation data. The learned action space enables direct latent-space planning and control for zero-shot cross-embodiment and dexterous manipulation.}
    \label{fig:overview}
\end{strip}

\begin{abstract}
World models facilitate planning and control in robotic manipulation by enabling agents to anticipate the consequences of their actions. However, they typically rely on action-labelled interaction data that is costly to acquire and difficult to scale. These models are also often tied to the specific embodiment they were trained on, limiting transferability to previously unseen robot morphologies. We introduce a 3D latent-action world model that learns continuous action representations directly from point-cloud observations of retargeted human manipulation demonstrations. A contrastive objective is used to encourage consistent representations of motion across embodiments. The resulting unified latent action space serves directly as the search space for model-predictive control, where sampled latent actions condition particle-based future predictions. Executable robot commands are then recovered from the predicted geometric motion without requiring an embodiment-specific action decoder. Through extensive evaluations, we demonstrate that the learned latent actions exhibit controllable and compositional motion, transfer to unseen robot embodiments, and support zero-shot manipulation with unseen gripper geometries. Following task-agnostic adaptation, our model achieves 90\%, 92\%, and 60\% success on reaching, grasping, and placing, respectively, compared with DexWM's reported 72\%, 58\%, and 28\%. Finally, we demonstrate zero-shot real-world manipulation with an unseen dexterous hand, achieving an 86\% success rate across 50 trials on five diverse tasks involving pushing, grasping, and lifting everyday objects. These results highlight the potential of geometry-grounded latent actions as a unified, transferable interface for world-model-based robotic manipulation. Project materials, including demonstration videos and source code, are available on our \href{http://vinaymohan0409.github.io/DexLatentWorld/}{project page.}

\end{abstract}

% \begin{itemize}
%     \item \textbf{Problem:} Action-conditioned robotic world models typically rely on expensive, embodiment-specific action labels.
%     \item \textbf{Gap:} Action-unlabelled interaction observations contain motion information, but inferred latent actions must be controllable and transferable across morphologies.
%     \item \textbf{Method:} Joint 3D particle-based inverse/forward world model, continuous latent actions, and contrastive alignment of retargeted cross-embodiment transitions.
%     \item \textbf{Control:} Plan with CEM directly over the learned latents; recover executable robot targets from predicted particle geometry without a learned latent-to-action decoder.
%     \item \textbf{Main evidence:} Unseen-gripper Push Puck 5/10, Push Cube 8/10, Transport/Place 10/10, Stacking 10/10; dexterous Reach/Grasp/Place 100/90/54\% vs. DexWM 72/58/28\%.
%     \item \textbf{Critical qualification:} The dexterous results follow task-agnostic robot-domain adaptation; the gripper experiments use no target-embodiment world-model adaptation.
%     \item \textbf{Takeaway:} Observation-derived 3D action latents unify transferable prediction and model-based robotic control.
% \end{itemize}    
% \newpage
\section{Introduction}
\label{sec:intro}

Developing world models that generalise across diverse robot embodiments remains a central challenge in robotic manipulation. Embodiments differ substantially in their morphologies, kinematics, and action spaces, making it difficult to learn predictive dynamics that transfer across platforms. This motivates learning a unified action representation that enables a world model to capture reusable interaction dynamics across different embodiments, including those unseen during training.

A promising direction is to learn action representations directly from observed manipulation, without relying on embodiment-specific action labels \cite{Ye_2025_ICLR}. Human manipulation videos provide a rich source of such observations, but representations learned from human motion alone may not generalise across robot morphologies. Retargeting these demonstrations to different robot embodiments produces corresponding manipulation trajectories that preserve the underlying interaction despite differences in morphology \cite{Zhang_2026_CVPR}. These correspondences provide a natural basis for learning a shared latent action space that captures motion rather than embodiment-specific variation.

Manipulation involves physical interactions that evolve over space and time. Explicit 3D geometry provides a natural way to represent this evolution, capturing the spatial relationships and contact geometry between interacting entities \cite{huang2026pointworld,Ai_2025_SciRob}. Representing scene evolution through particle displacements provides direct supervision of physical motion, allowing predictive dynamics to focus on how the scene evolves rather than its visual appearance. We therefore seek to ground both latent action inference and predictive dynamics in a common geometric representation, enabling actions to be learned and their consequences to be predicted in the same physical space.

% Beyond predicting interaction dynamics, a learned latent action space also offers a structured representation for model-based planning. This is particularly relevant to dexterous manipulation, where searching over high-dimensional joint-action spaces can be computationally demanding \cite{goswami2026dexwm}. \textcolor{red}{By representing scene evolution through particle displacements, we can model changes in physical geometry directly. We therefore seek to ground both latent action inference and predictive dynamics in a common 3D representation, linking observed motion to its predicted consequences }\cite{Yamada_2025_CoRL, yang2026lamp}.

In this work, we introduce DexLatentWorld, a unified 3D latent-action world model for prediction and control, learning continuous action representations directly from action-unlabelled manipulation observations (Fig. \ref{fig:overview}). We represent these observations as temporally corresponding 3D scene particles derived from tracked hand–object trajectories, with human manipulation trajectories retargeted to different robot embodiments. A cross-embodiment contrastive objective aligns corresponding human and robot transitions within a shared latent action space, while a forward dynamics model uses these representations to predict scene evolution in 3D particle space. At deployment, we perform closed-loop model-predictive control (MPC) directly in the learned latent action space, evaluating candidate latent-action sequences through their predicted geometric consequences. We then recover executable robot commands from predicted motion using known robot geometry and kinematics, without additional action-labelled demonstrations for training an embodiment-specific latent-to-action decoder.

The learned latent actions exhibit controllable and compositional motion, enabling consistent geometric predictions across unseen robot embodiments. To evaluate the effectiveness of these capabilities in robotic manipulation, we employ MPC in both simulation and real-world deployment. We further demonstrate zero-shot manipulation with unseen gripper geometries without embodiment-specific adaptation. Extending to dexterous manipulation, our framework achieves 90\%, 92\%, and 60\% success on reaching, grasping, and placing following task-agnostic domain adaptation, compared with DexWM \cite{goswami2026dexwm}'s 72\%, 58\%, and 28\%. In real-world deployment, the original pretrained model achieves 86\% success on five manipulation tasks with an unseen dexterous hand, without any fine-tuning. Our contributions include:
\begin{itemize}
    \item \textbf{A unified 3D latent-action world model.}
    We introduce a self-supervised framework that combines
    particle-based scene dynamics with a shared continuous
    latent action representation, enabling cross-embodiment
    world modelling and direct latent-space planning and control.
    
    \item \textbf{A geometry-based prediction-to-control framework.} We plan directly in the learned latent action space and recover executable robot targets from predicted 3D motion, avoiding the need for a learned embodiment-specific latent-to-action decoder.

    \item \textbf{Cross-embodiment generalisation and dexterous manipulation.}  We demonstrate zero-shot transfer to held-out end effectors, including real-world dexterous manipulation without world-model adaptation, and task generalisation following task-agnostic domain adaptation.
\end{itemize}

\section{Related Work}
\label{sec:related_work}

\vspace{0.05cm}
\noindent\textbf{Latent Action Models.}
Learning latent actions from observations has emerged as a promising approach to leveraging large-scale, unlabelled interaction data. Typically, an inverse dynamics model encodes changes between observations into a latent action, which is then used to condition a forward dynamics model to predict the resulting scene evolution~\cite{Schmidt_2024_ICLR,Ye_2025_ICLR}. However, the learned representation does not necessarily isolate the underlying physical motion. When inferred from images, latent actions may become entangled with viewpoint changes, camera motion, background appearance, and other visual variations. Reconstruction-based objectives can also encourage shortcut learning, allowing latent representations to encode information about future observations rather than meaningful action-like transformations. To address these challenges, prior work has explored approaches including action-category-guided contrastive learning~\cite{Dai_2026_arXiv}, early temporal differencing and temporal contrastive learning~\cite{Yang_2026_CVPR}, cross-viewpoint reconstruction~\cite{Lee_2026_ICML}, and contrastive alignment of video-derived latent actions with executable action tokens learned from robot trajectories~\cite{Zhang_2026_arXiv}. We pursue a complementary direction by inferring latent actions directly from 3D point representations of the scene and aligning matched manipulation transitions across embodiments, encouraging shared motion representations across different morphologies.

\vspace{0.05cm}
\noindent\textbf{World Models for Robotic Manipulation.}
Learning predictive dynamics has become an important direction in robotic manipulation, supporting model-based planning and control~\cite{Hou_2026_IJRR}. Much of this research has explored video-based world models, with more recent approaches investigating explicit 3D representations to model spatial interactions and scene evolution. PointWorld~\cite{huang2026pointworld} scales particle-based world modelling to diverse manipulation settings, predicting full-scene 3D point displacements through a shared geometric state and action representation. 3DPWM~\cite{peri2026pwm} addresses partial observability through point-cloud completion, enabling dynamics prediction over temporally corresponding 3D geometry. Exploring this direction in dexterous manipulation, He et al.~\cite{he2026scaling} train a graph-based particle dynamics model across different hand embodiments, demonstrating generalisation to unseen morphologies. These works demonstrate the effectiveness of explicit 3D representations for predictive modelling in robotic manipulation.

\vspace{0.05cm}
\noindent\textbf{Latent Action World Models.}
Early work demonstrated that latent actions inferred from unlabelled videos could provide a conditioning interface for interactive world modelling~\cite{Bruce_2024_ICML,Gao_2025_ICML}. These approaches typically learn the latent action representation separately from the predictive model, limiting their ability to adapt jointly. This motivates joint training to better align action representations with predictive dynamics~\cite{Wang_2026_ICML,Garrido_2026_ICML,Ayalew_2026_arXiv}. More recently, LAC-WM~\cite{huang2026lacwm} extends latent-action world modelling across human and robot embodiments, learning a unified action representation using action-labelled motion supervision during training. Adapting the model to unseen robots additionally requires labelled robot interactions to train an action-to-latent projector. LAWM-3D~\cite{Yang_2026_arXiv_a} introduces geometric constraints and multi-view consistency to learn 3D-aware latent actions; however, its world model remains video-based. Despite these advances, planning for dexterous manipulation remains challenging, with existing approaches employing large sampling budgets and task-specific initialisation~\cite{goswami2026dexwm}, manually designed motion primitives~\cite{he2026scaling}, or learned motion priors~\cite{Yamada_2025_CoRL,yang2026lamp} to facilitate planning and control. In contrast, we optimise directly in the latent action space, recovering executable robot targets from predicted particle motion through known geometry and kinematics.
\section{Method}
\label{sec: method}

\begin{figure*}
    \centering
    \includegraphics[width=1\linewidth]{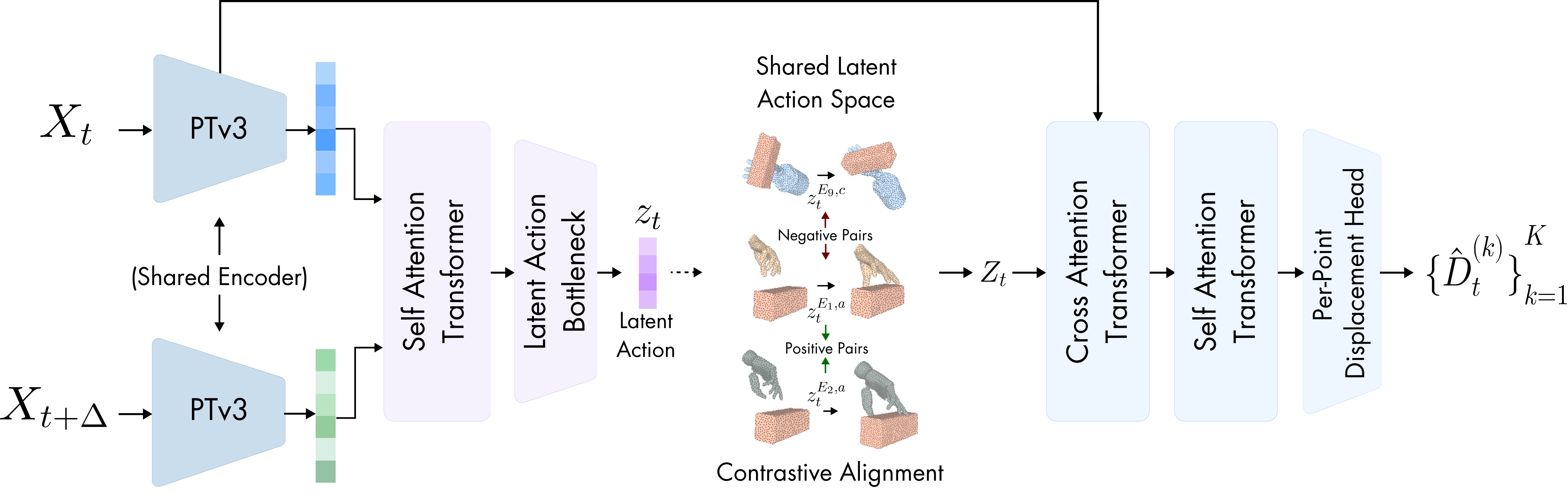}
    \caption{
    \textbf{Overview of DexLatentWorld.}
     An inverse dynamics model encodes scene transitions into
    compact latent actions, which are aligned across embodiments through
    contrastive learning. Here, $E_i$ denotes an embodiment or morphology
    (e.g., human, Allegro, XHand), while $a$, $b$, $c$, $\ldots$
    denote different transitions. $z_t^{E_1,b}$ represents the latent action inferred from transition
    $b$ performed by embodiment $E_1$. The forward dynamics model conditions on the current scene
    and a latent action sequence $Z_t$, comprising five latent actions
    in our experiments, to predict per-point displacement fields
    $\{\hat{D}_t^{(k)}\}_{k=1}^{K}$, where $\hat{D}_t^{(k)}$
    denotes the predicted displacement field at horizon $k$.
    }
    \label{fig:placeholder}
\end{figure*}

\subsection{Problem Formulation}
% \begin{itemize}
%     \item H2O data
%     \item Contact refinement - quantitative measurements
%     \item Unidex style retargetting
%     \item Barycentric Sampling
%     \item Table and Object Augmentation
% \end{itemize}

We formulate 3D world modelling as latent-action-conditioned prediction of temporally corresponding scene particles. Each state $X_t \in \mathbb{R}^{N \times 3}$ represents the human or robot hands, manipulated object, distractors, and background geometry. During training, an inverse dynamics model infers a latent action from each consecutive pair of observed states, $z_t = q_{\phi}(X_t, X_{t+\Delta})$. Given the initial scene $X_t$ and a sequence of $K$ latent actions, the forward dynamics model jointly predicts the next $K$ states in a single forward pass. Prediction at horizon $k$ is conditioned only on the initial scene and the first $k$ latent actions, $\hat{X}_{t+k\Delta} = f_{\theta}(X_t, z_t, \ldots, z_{t+(k-1)\Delta})$, using causal chunked decoding. We use $K=5$ and sample states every six frames from H2O's 30\,Hz data, giving $\Delta=0.2$\,s and a prediction horizon of 1\,s \cite{kwon2021h2o}. This interval is chosen to capture appreciable scene motion while retaining fine-grained motion.

\subsection{Latent Action World Model}

\vspace{0.1cm}
\noindent\textbf{Shared Scene Encoding.}
We jointly learn inverse and forward dynamics models using a shared Point Transformer V3 (PTv3) backbone \cite{Wu_2024_CVPR}. We adopt Utonia's PTv3 implementation~\cite{zhang2026utonia} and train the backbone from scratch. The encoder produces a four-stage hierarchy of geometric features, combining local surface detail with progressively broader spatial context. The inverse dynamics model uses attention between the coarsest PTv3 features extracted from consecutive observations to infer the transition, whereas the forward dynamics model leverages multi-scale features from the initial scene to predict particle motion.

\vspace{0.1cm}
\noindent\textbf{Latent Action Inference.} For each consecutive pair of observations $(X_t, X_{t+\Delta})$, we augment their coarsest PTv3 features with temporal embeddings encoding their observation order. We then concatenate the two feature sequences with a separator token and four learnable query tokens, allowing a self-attention transformer to capture the underlying motion between observations. Each query output is projected to eight dimensions, yielding a continuous latent action $z_t \in \mathbb{R}^{4 \times 8}$. This 32-dimensional bottleneck encourages a compact representation of the observed transition, while retaining multiple tokens allows individual particles to attend selectively to different aspects of the inferred action. The latent representation is learned jointly through prediction and cross-embodiment alignment objectives.

\vspace{0.1cm}
\noindent\textbf{Causal Multi-Horizon Prediction.} Rather than directly predicting future particle positions, we formulate the forward dynamics model to predict displacements relative to the initial scene. This encourages the model to focus on temporal changes in particle positions rather than learning to reproduce static scene information. Given the initial observation $X_t$ and a sequence of $K$ inferred latent actions, the decoder predicts particle-level displacements across all future horizons. To obtain particle-level representations for predicting each input particle's displacement, we map features from all PTv3 encoder stages back to the finest resolution using the retained pooling indices. The resulting features are concatenated, projected into a shared 192-dimensional space, and replicated across \(K\) horizons as separate batch elements. Cross-attention and self-attention operate independently within each horizon, preventing information exchange across horizons. Each particle then queries the projected latent-action tokens through cross-attention, with learned step embeddings encoding their temporal positions. The action-conditioned features are further processed by an AdaLN self-attention block, enabling information exchange across particles to capture broader scene context. To prevent future-action leakage within each prediction horizon, later action tokens are masked in cross-attention and their corresponding entries are zeroed in the AdaLN conditioning vector, ensuring that prediction at horizon $k$ depends only on the initial scene and the first $k$ latent actions. A point-wise output head predicts cumulative displacements, which are added to the initial particle positions to obtain the future scene, $\hat{x}_{t+k\Delta,i} = x_{t,i} + \hat{d}^{(k)}_{t,i}$, for $k=1,\ldots,K$. By sharing the initial scene representation across horizons, all $K$ future states are predicted in a single batched decoder pass without autoregressive feedback.

\subsection{Training Objective}

We train the shared PTv3 encoder, inverse dynamics model, and forward dynamics model end-to-end using
\begin{equation}
    \mathcal{L}
    = \mathcal{L}_{\mathrm{disp}}
    + \lambda_{\mathrm{con}}\mathcal{L}_{\mathrm{con}},
\end{equation}
where \(\lambda_{\mathrm{con}}=0.15\) weights the contrastive loss. The displacement loss requires latent actions to retain predictive information, while the contrastive loss acts as a regulariser encouraging corresponding transitions across embodiments to share a common representation.

\vspace{0.1cm}
\noindent\textbf{Movement-Weighted Displacement Loss.}
Only a small subset of scene particles moves during manipulation, so a uniformly weighted loss is dominated by stationary regions. Following PointWorld~\cite{huang2026pointworld}, we use a movement-weighted Huber loss, weighting each particle at each prediction horizon by a soft movement score computed from its ground-truth displacement. For particle $i$ at horizon $k$, the ground-truth displacement is defined as $d_{t,i}^{(k)} = x_{t+k\Delta,i} - x_{t,i}$. We then compute
{\small
\begin{equation}
\begin{aligned}
m_{t,k,i} &= \sigma\left(\kappa\left(\|d_{t,i}^{(k)}\|_2-\tau_m\right)\right),\\
w_{t,k,i} &= \frac{m_{t,k,i}}{\sum_{j=1}^{N}m_{t,k,j}},
\end{aligned}
\end{equation}
}
where $\sigma$ is the sigmoid function, $\tau_m$ controls the movement threshold, and $\kappa$ controls the sharpness of the transition. This soft weighting emphasises moving regions while retaining supervision over the full scene. We normalise displacements using training-set statistics computed separately for each spatial dimension and prediction horizon, then apply a Huber loss:
\begin{equation}
    \mathcal{L}_{\mathrm{disp}}
    = \frac{1}{K}
      \sum_{k=1}^{K}\sum_{i=1}^{N}
      w_{t,k,i}\,
      \operatorname{Huber}\!\left(
          \hat{d}_{t,i}^{(k),\mathrm{norm}}
          - d_{t,i}^{(k),\mathrm{norm}}
      \right).
\end{equation}
Normalisation balances displacement scales across prediction horizons and spatial dimensions, while the Huber loss reduces sensitivity to large residuals.

\vspace{0.1cm}
\noindent\textbf{Cross-Embodiment Alignment.}
Since $X_t$ already provides scene and embodiment geometry, we use a contrastive objective to encourage $z_t$ to capture transferable motion by aligning latent actions from corresponding transitions across embodiments \cite{NEURIPS2020_d89a66c7}. Each training window is sampled under two randomly selected embodiments, whose first-transition latents $z_t$ form a positive pair. Other windows in the mini-batch serve as negatives, with at most one window per sequence to avoid pushing similar motions apart. For each anchor transition $a\in\mathcal{A}$, we flatten its latent action into a vector $z_a$ and compute temperature-scaled cosine similarity $s_{ab} = \cos(z_a,z_b)/\tau_c$, where $\tau_c$ is the contrastive temperature. Denoting the positive set by $\mathcal{P}(a)$ and the set of valid comparisons, excluding the anchor itself, by $\mathcal{V}(a)$, the contrastive loss is
\begin{equation}
    \mathcal{L}_{\mathrm{con}}
    = -\frac{1}{|\mathcal{A}|}
      \sum_{a\in\mathcal{A}}
      \frac{1}{|\mathcal{P}(a)|}
      \sum_{b\in\mathcal{P}(a)}
      \log
      \frac{\exp(s_{ab})}
           {\sum_{c\in\mathcal{V}(a)}\exp(s_{ac})},
\end{equation}
where $c$ indexes comparison transitions. Jointly optimising displacement prediction and contrastive alignment encourages a representation that preserves predictive motion information while reducing embodiment-specific variation.

\subsection{Latent-Space Planning and Control}

At deployment, we use the learned dynamics model for goal-directed planning by optimising latent-action sequences directly. Given the current scene $X_t$, we search for a sequence $Z_t=(z_t,\ldots,z_{t+(K-1)\Delta})$ whose predicted scene evolution minimises a task-specific geometric cost:
\begin{equation}
Z_t^* = \arg\min_{Z_t} C(f_\theta(X_t,Z_t)).
\end{equation}
We solve this optimisation using the Cross-Entropy Method (CEM)~\cite{deboer2005cem}, initialised with a diagonal Gaussian fitted to latent actions inferred from training transitions. For fine-tuned models, these statistics are estimated from their task-agnostic adaptation sequences. CEM samples candidate latent-action sequences, evaluates their predicted particle trajectories against the task objective, and iteratively refits the sampling distribution to the elite candidates. We adopt a receding-horizon strategy~\cite{mayne2000mpc}, repeatedly replanning from the latest observation and executing the initial portion of each optimised trajectory.

To execute the predicted motions, we recover robot configurations from the corresponding particles using known geometry and kinematics. For grippers, predicted particle centroids provide Cartesian position targets, while rigid alignment can additionally recover target orientations. For dexterous hands, we estimate the palm pose through Kabsch alignment~\cite{kabsch1976}, exploiting the known particle correspondences, and subsequently refine finger joint targets through bounded nonlinear least-squares inverse kinematics. The resulting targets are passed to embodiment-specific low-level controllers, enabling robot control without a learned latent-to-action decoder.

\section{Dataset Curation \& Experimental Protocol}
\label{sec: data_preprocessing}

We construct training sequences from a subset of H2O \cite{kwon2021h2o} bimanual manipulation data, using fitted MANO \cite{romero2017mano} hand meshes, object meshes, and tracked object poses. Training and validation sets are partitioned at the sequence level to prevent temporal overlap. We first refine the human hand trajectories to reduce spurious hand--object separation and excessive interpenetration while preserving the demonstrated motion and keeping object trajectories fixed. Refinement reduces contact-region Chamfer distance by 15.1\% and the median 95th-percentile penetration depth across hand trajectories by 22.6\%. We then retarget the human hand motion to eight robotic hands using UniDex-style wrist alignment and fingertip inverse kinematics, yielding matched transitions across one human and eight robot embodiments \cite{Zhang_2026_CVPR}. We sample 1,024 particles per entity, including each hand, manipulated object, table, and static distractors, using area-weighted barycentric sampling where mesh geometry is available. We additionally reconstruct the table by fitting a plane to unprojected multi-frame depth observations using RANSAC. Static distractor objects are placed once per sequence outside an expanded bounding box enclosing the hand and manipulated object over the full trajectory, varying scene clutter across embodiments while preserving the underlying interaction.

We evaluate geometric prediction accuracy using mover endpoint error (Mover EPE), defined as the mean per-point Euclidean distance between predicted and ground-truth 3D positions, measured in millimetres. Hand particles outnumber object particles 2:1 in each scene. Since a larger proportion of hand particles move more than 5 mm, they account for 74.0\% of moving particles, compared to 26.0\% for object particles, pooled across all five prediction horizons. We therefore additionally report hand- and object-specific EPE, alongside EPE for static scene points (table and distractors). We focus on moving particles to prevent stationary scene geometry from obscuring prediction errors in regions undergoing manipulation. Moving particles are identified by a ground-truth displacement exceeding 5 mm at each prediction horizon. We evaluate predictions over five steps spanning 1 s, with errors reported at the horizons specified in each experiment. This dense geometric metric captures systematic differences in rollout fidelity that task-level success rates may not reveal. Since predictions span only one second, absolute differences in EPE can appear modest despite meaningful differences in predicted motion. We therefore assess geometric fidelity alongside qualitative rollouts and downstream task success rates.

\section{Experiments}
\label{sec:experiments}

We investigate whether latent actions learned from action-unlabelled 3D observations support transferable prediction and control across robot end effectors. We examine latent conditioning, composition, and cross-embodiment transfer (Sections \ref{sub:conditioning_ablation}--\ref{sub:generalise}). We then evaluate zero-shot gripper manipulation (Section \ref{sub:gripper}), planning-space and execution-strategy ablations (Section \ref{sub:planning}), and dexterous task generalisation following task-agnostic adaptation (Section \ref{sub:dex}). Finally, we assess real-world deployment on an unseen dexterous hand without fine-tuning (Section \ref{sec:real_world}).

% We  whether latent actions learned from action-unlabelled 3D manipulation observations provide a shared representation for prediction and control across robot embodiments. Our evaluation addresses four questions. First, does the world model rely on transition-specific latent actions to predict scene evolution (Section \ref{sub:conditioning_ablation})? Second, does the learned action space support interpolation, composition, and reuse across different initial configurations (Section \ref{sub:express})? Third, can latent actions inferred from one embodiment transfer to others, including robot hands unseen during training (Section \ref{sub:generalise}). Finally, can the learned latent space serve directly as a planning interface, with executable robot commands recovered from predicted geometry rather than a learned action decoder (Section \ref{sub:planning})? We examine these properties through quantitative prediction errors, controlled latent-action interventions, geometric rollouts, and task success. We further evaluate zero-shot manipulation with an unseen parallel-jaw gripper (Section \ref{sub:gripper}) and generalisation to unseen dexterous manipulation tasks following task-agnostic world-model adaptation (Section \ref{sub:dex}).

\subsection{Latent Conditioning Ablations}\label{sub:conditioning_ablation}

Given an observed manipulation window, we infer its latent action sequence and condition the forward dynamics model on the initial scene to predict subsequent states. To assess whether the model relies on latent conditioning, we compare predictions using the inferred latents against three interventions: zero conditioning, latent sequences shuffled across unrelated trajectories, and a fixed latent sequence shared across evaluation windows. As shown in Table~\ref{tab:latent_ablation}, correct conditioning achieves mean hand and object EPEs of 7.94\,mm and 13.59\,mm, respectively, substantially outperforming zero (53.17/43.43\,mm), shuffled (75.01/51.96\,mm), and fixed (60.29/46.78\,mm) conditioning. On average, static-scene errors remain below 1\,mm across all latent conditions. These results demonstrate that trajectory-aligned latent actions are essential for accurately predicting hand and object motion.

% \begin{table}
% \centering
% \footnotesize
% \setlength{\tabcolsep}{4pt}
% \begin{tabular}{@{}lcccccc@{}}
% \toprule
%  & \multicolumn{5}{c}{Rollout step} & \\
% \cmidrule(lr){2-6}
% Prediction & 1 & 2 & 3 & 4 & 5 & Mean \\
% \midrule
% No motion (baseline) & 25.33 & 40.19 & 52.46 & 63.15 & 72.53 & 50.73 \\
% \midrule
% Correct latents & \textbf{5.52} & \textbf{7.77} & \textbf{9.61} & \textbf{11.31} & \textbf{12.88} & \textbf{9.42} \\
% Zero latents & 25.26 & 40.09 & 52.31 & 62.90 & 72.36 & 50.58 \\
% Shuffled latents & 32.94 & 54.93 & 72.35 & 86.38 & 97.83 & 68.89 \\
% Fixed latents & 28.98 & 46.64 & 59.67 & 69.69 & 78.68 & 56.73 \\
% \bottomrule
% \end{tabular}
% \caption{\textbf{Latent actions carry the motion.} Mean Mover EPE in mm ($\downarrow$) of the predicted cumulative displacement, over 4{,}000 validation windows. "No motion" assumes zero displacement
% for every point. The other rows decode with the model's own latents, latents set to zero, latents taken from other sequences (shuffled), or latents taken from one fixed window (fixed donor).}
% \label{tab:latent_ablation}
% \end{table}

\begin{table}[t]
\centering
\footnotesize
\setlength{\tabcolsep}{2pt}
\resizebox{\columnwidth}{!}{%
\begin{tabular}{@{}l*{9}{c}@{}}
\toprule
& \multicolumn{3}{c}{Step 1}
& \multicolumn{3}{c}{Step 5}
& \multicolumn{3}{c}{Mean} \\
\cmidrule(lr){2-4}
\cmidrule(lr){5-7}
\cmidrule(lr){8-10}
Prediction
& Hand & Object & Scene
& Hand & Object & Scene
& Hand & Object & Scene \\
\midrule
No motion (baseline)
& 25.97 & 23.18 & 0.00
& 76.78 & 61.50 & 0.00
& 53.28 & 43.70 & 0.00 \\
\midrule
Correct latents
& \textbf{4.61} & \textbf{8.51} & 0.39
& \textbf{10.87} & \textbf{18.12} & 1.34
& \textbf{7.94} & \textbf{13.59} & 0.87 \\
Zero latents
& 25.94 & 23.03 & 0.28
& 76.66 & 61.18 & 0.84
& 53.17 & 43.43 & 0.68 \\
Shuffled latents
& 34.90 & 26.49 & 0.39
& 107.72 & 72.29 & 1.21
& 75.01 & 51.96 & 0.82 \\
Fixed latents
& 30.34 & 24.33 & 0.34
& 84.06 & 64.81 & 0.94
& 60.29 & 46.78 & 0.70 \\
\bottomrule
\end{tabular}%
}
\caption{\textbf{Latent conditioning drives motion prediction.}
Particle EPE in mm ($\downarrow$) at steps 1 and 5, with the mean computed across all five prediction horizons on our H2O validation split (nine training embodiments).
Hand and object errors are evaluated on points with ground-truth displacement $> 5$ mm, while scene errors include static table and distractor points.
No motion assumes zero displacement.  Other rows
use correct, zeroed, shuffled, or fixed-donor latent actions.}
\label{tab:latent_ablation}
\end{table}

\subsection{Latent Action Structure and Composition}\label{sub:express}

\begin{figure}
    \centering
    \includegraphics[width=0.95\columnwidth]{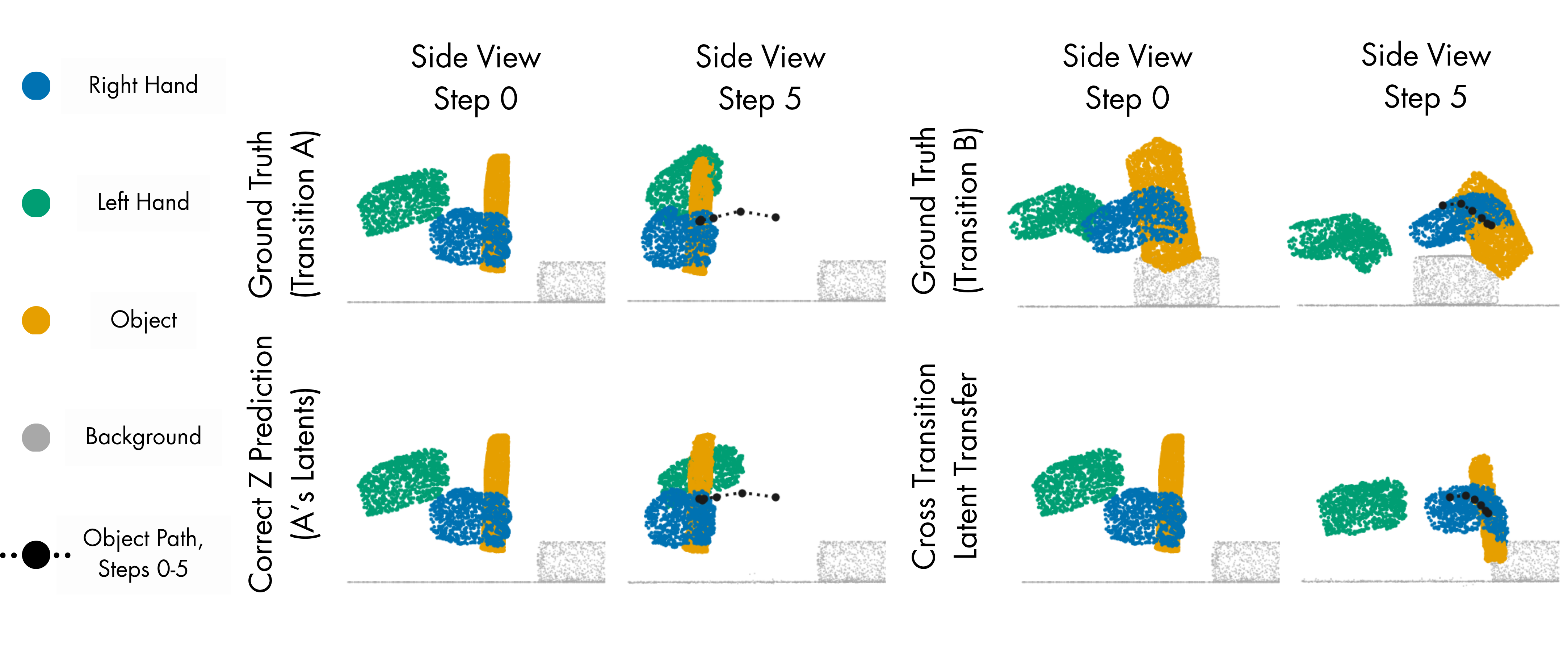}
    \caption{\textbf{Counterfactual evaluation.}
     Left: Latent actions inferred from the ground-truth sequence of Transition A are applied to its initial frame. The world model's predictions closely follow the ground-truth trajectories. Right: Latent actions inferred from Transition B are applied to the initial frame of Transition A, inducing a corresponding motion pattern. Results are shown at steps 0 and 5, with dotted lines indicating object trajectories.}
    \label{fig:counterfactual}
\end{figure}

\begin{figure}
    \centering
    \includegraphics[width=0.8\linewidth]{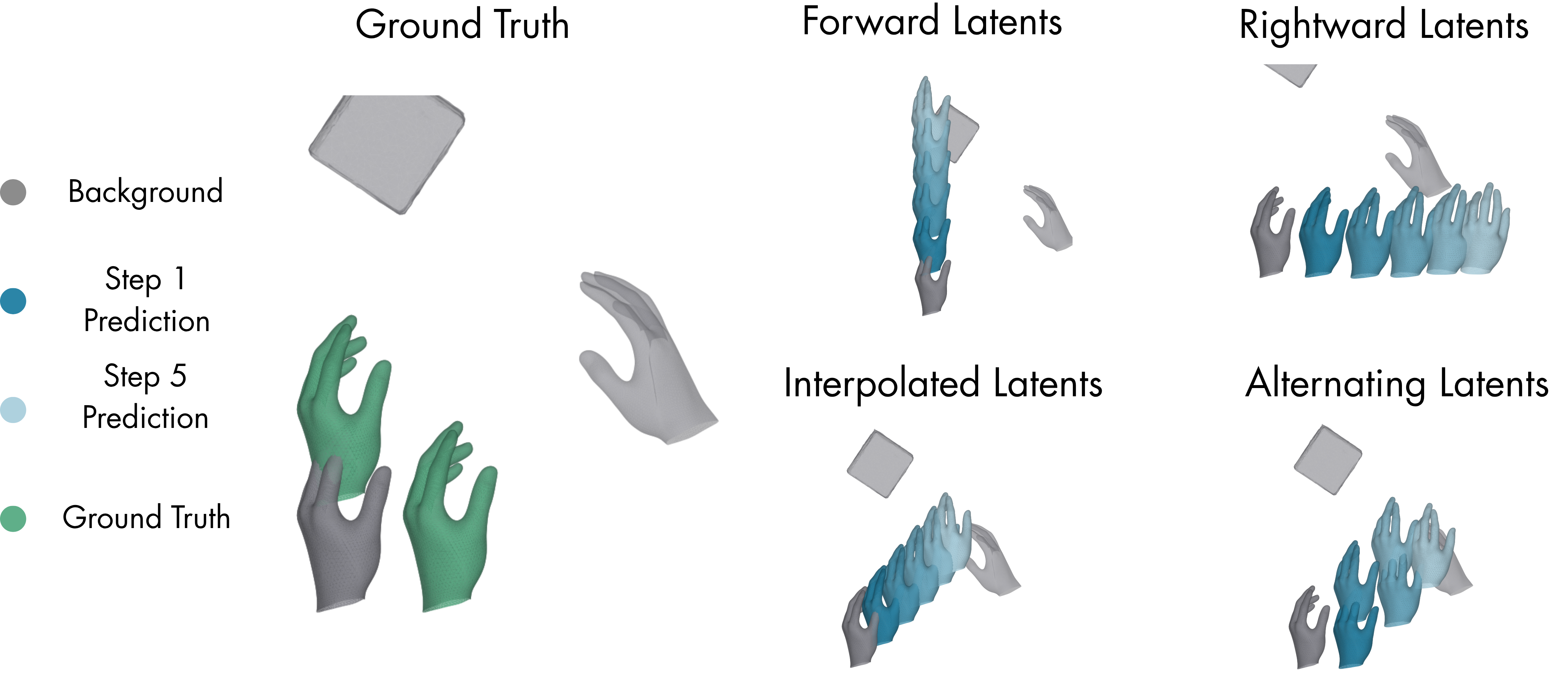}
    \caption{\textbf{Action composition. }
    Latents inferred from synthetic forward and rightward hand motions are repeated, averaged, and alternated to visualise their effects on predicted trajectories.}
    \label{fig:latent_composability}
\end{figure}

Applying latents inferred from one transition to a different initial scene produces coherent counterfactual motion rather than merely degrading prediction quality (Fig.~\ref{fig:counterfactual}). This demonstrates that latent actions can induce meaningful motion beyond the transitions from which they were inferred. We also investigate interpolation and composition using latents inferred from synthetically generated forward and rightward hand motions. Repeating each latent produces consistent motion in its corresponding direction, while averaging the latents yields an intermediate trajectory and alternating them produces composed motion (Fig.~\ref{fig:latent_composability}). These results illustrate the reusability and compositional structure of the learned action space.

\subsection{Cross-Embodiment Generalisation}\label{sub:generalise}

\begin{figure}
    \centering
    \includegraphics[width=0.9\linewidth]{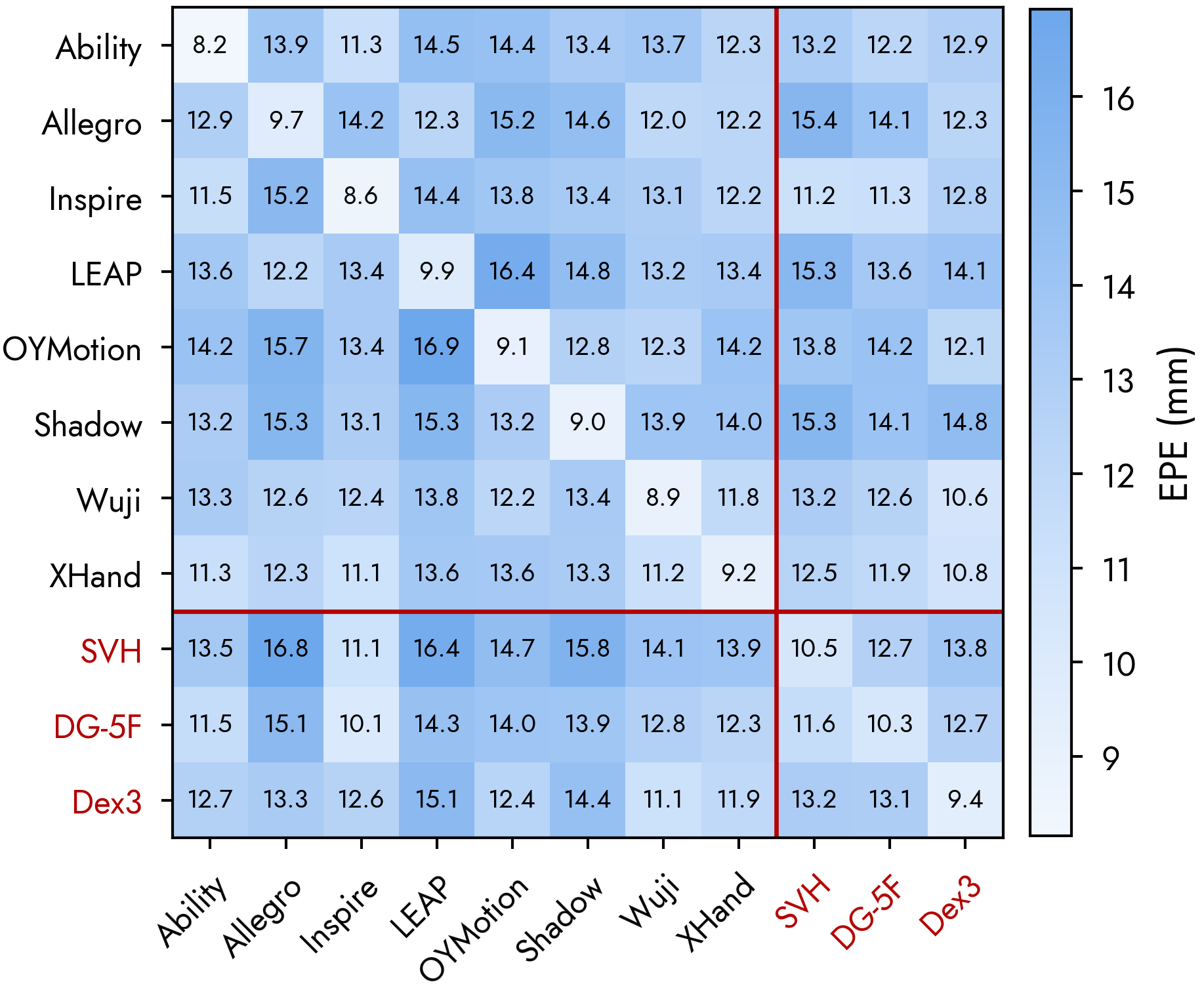}
    \caption{\textbf{Cross-embodiment latent transfer.} Each cell reports the Mover EPE averaged over five rollout steps when latent actions inferred from the source embodiment (rows) are decoded using the initial point cloud of the target embodiment (columns). All evaluations use the same 256 validation windows. Embodiments highlighted in red (SVH, DG-5F, Dex3) were unseen during training.}
    \label{fig:cross_decode}
\end{figure}

\begin{figure}
    \centering \includegraphics[width=1\linewidth]{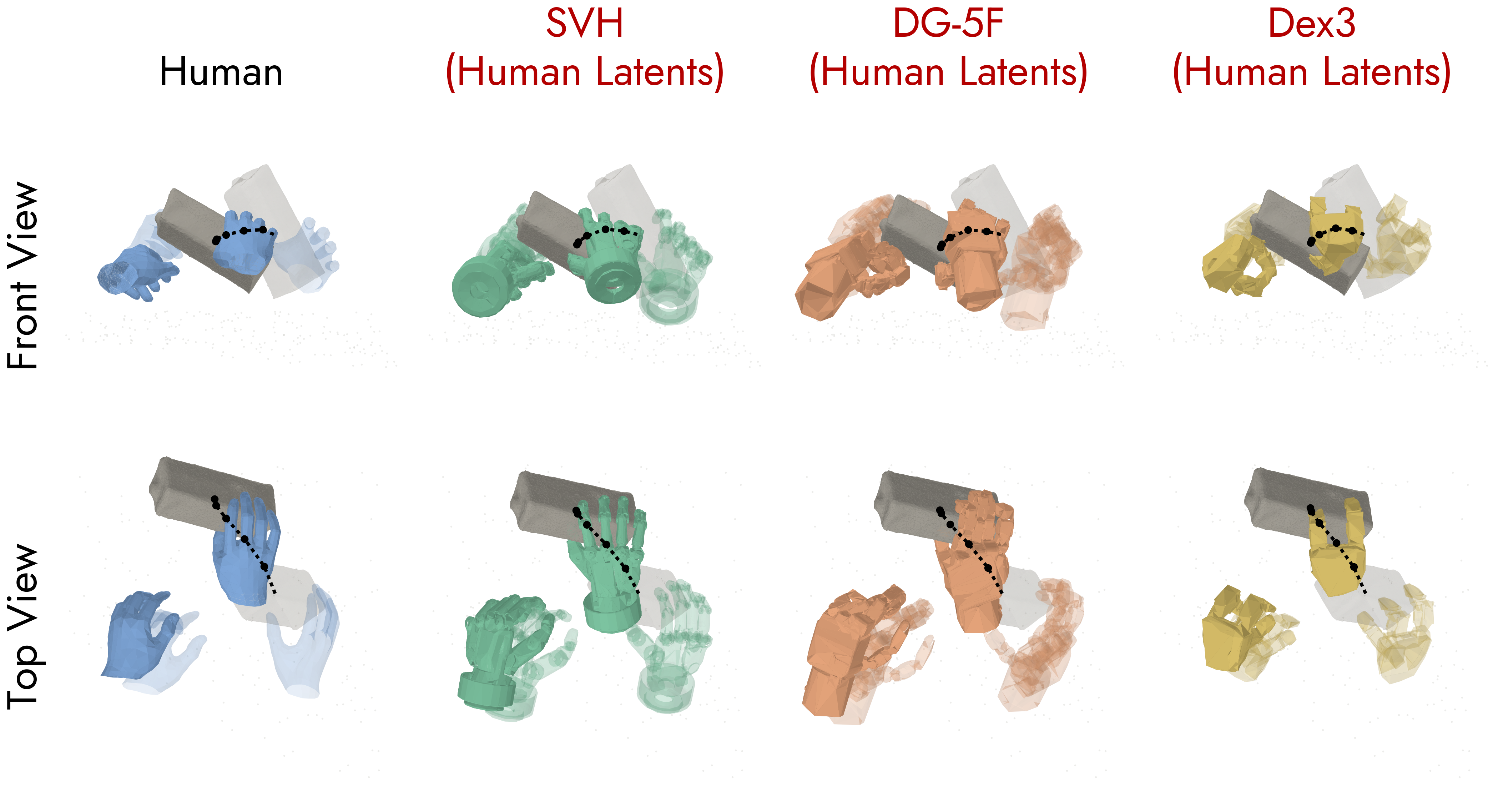}
    \vspace{-0.8cm}
    \caption{\textbf{Human-to-robot transfer.} Latent actions are encoded from a human pouring demonstration (our validation set) and decoded with the initial point cloud of each held-out hand (SVH, DG-5F, Dex3). The dotted line traces the carton's centre over steps 1–5. For clearer visualisation, predicted points are rendered as meshes to better illustrate the resulting motion.}
    \label{fig:emb_transfer}
\end{figure}

We evaluate whether the learned latent actions capture motion that transfers across different robot morphologies, including three embodiments unseen during world-model training: SVH, DG-5F, and Dex3. We first assess cross-embodiment transfer through cross-decoding. Latent actions are inferred from a source manipulation trajectory and used to predict scene evolution from the initial point cloud of the corresponding retargeted trajectory in a target embodiment. The resulting predictions are compared against the ground-truth retargeted trajectory using Mover EPE. As shown in Fig.~\ref{fig:cross_decode}, cross-decoding errors across different embodiments are broadly comparable to same-embodiment predictions and remain substantially below the 49.5 mm pooled no-motion baseline. We further illustrate this transfer qualitatively in Fig.~\ref{fig:emb_transfer}, where latent actions inferred from a human pouring demonstration condition open-loop predictions for the three unseen robot hands without additional world-model adaptation. Finally, we assess embodiment-specific information in the learned representations using a linear classification probe. Embodiment classification accuracy is 26.6\% for latent actions, compared with 85.3\% for pooled final-stage PTv3 encoder features (9.1\% chance across 11 robot embodiments), indicating substantially reduced embodiment decodability while preserving motion information that transfers across different morphologies.

\subsection{Zero-Shot Gripper Manipulation}\label{sub:gripper}

We investigate zero-shot transfer to gripper geometries absent from pretraining, using the original H2O-pretrained checkpoint without fine-tuning. Using a Franka Panda in RoboSuite \cite{robosuite2020}, we evaluate a Robotiq140 parallel-jaw gripper on Cube Stacking and Milk Transport, and a WipingGripper on Push Cube and Reach + Push Puck. Stacking and transport begin with the object grasped. For Push Cube, the gripper starts behind the cube, whereas Reach + Push Puck requires the planner to position the gripper behind the puck before pushing. We optimise latent actions using CEM over point clouds sampled from simulator geometry, executing predicted gripper motion through operational-space control. Results in Table~\ref{tab:gripper_results} demonstrate that latent actions support manipulation across unseen gripper geometries without embodiment-specific action supervision.

\begin{table}[t]
\centering
\footnotesize
\setlength{\tabcolsep}{3pt}
\begin{tabular}{@{}lcccc@{}}
\toprule
Approach &
\shortstack{Reach +\\Push Puck} &
\shortstack{Push\\Cube} &
\shortstack{Milk\\Transport} &
\shortstack{Cube\\Stacking} \\
\midrule
PointWorld & -- & -- & 100\% & 100\% \\
Ours & 50\% & 80\% & 100\% & 100\% \\
\bottomrule
\end{tabular}
\caption{\textbf{Zero-shot gripper manipulation.}
Task success rates over 10 evaluation trials per task.
Our model is evaluated without world-model fine-tuning
on the target gripper. PointWorld results are reported
for the two shared manipulation tasks.}
\label{tab:gripper_results}
\end{table}

\subsection{Planning \& Execution Ablations}\label{sub:planning}

\vspace{0.1cm}
\noindent\textbf{World-Model Fine-Tuning.} We evaluate closed-loop manipulation in RoboCasa \cite{robocasa2024} under partial observability, using scene geometry reconstructed from two RGB-D views and forward-kinematics-derived hand geometry. Unlike the fully observed pretraining scenes, this setting introduces noisy, partial observations. To improve performance under this domain shift, we fine-tune the world model using rank-2 LoRA~\cite{hu2022lora} on 84 task-agnostic exploratory play sequences from 7 non-evaluation kitchens. Since these observations lack temporal particle correspondences, we use CoTracker3~\cite{Karaev_2025_ICCV} to estimate them for displacement supervision. To handle reconstruction noise, we extend the movement-weighted Huber loss with learned aleatoric uncertainty, replacing each weighted residual by $w_{t,k,i}(e^{-s_{t,k,i}}\ell_\delta+\lambda s_{t,k,i})$, where $\ell_\delta$ is the Huber residual, $s_{t,k,i}$ the predicted log-uncertainty, and $\lambda=1$.

\vspace{0.1cm}
\noindent\textbf{Planning and Execution.} We compare four configurations using the same fine-tuned world model, evaluation episodes, and planning budgets on Reach, Grasp, and Place (object initially grasped), with 20 trials per task. Action--IK and Action--Direct search native robot actions, mapped to latents by a two-layer MLP for world-model conditioning. The former recovers robot targets by fitting the kinematic hand model to predicted points, while the latter executes the optimised actions. Latent--IK (ours) searches in latent space and recovers robot targets from predicted hand motion. Latent--Decoder uses a trained two-layer latent-to-action MLP for execution. In Table~\ref{tab:planning_ablation}, Latent--IK matches native-action search on Reach (100\%) and Place (60\%) and reaches 90\% versus 85\% on Grasp. Although Latent--Decoder achieves higher success on Place (80\%), direct latent-space planning with geometry-based execution is competitive without requiring learned latent-to-action mappings.

\begin{table}[t]
\centering
\footnotesize
\renewcommand{\arraystretch}{1.15}
\begin{tabular*}{0.75\columnwidth}{
    @{\extracolsep{\fill}}lcccc@{}}
\toprule
Task
& $A_{\mathrm{IK}}$
& $A_{\mathrm{dir}}$
& $L_{\mathrm{IK}}$
& $L_{\mathrm{dec}}$ \\
\midrule
Reach & 100\% & 100\% & 100\% & 100\% \\
Grasp & 85\% & 85\% & 90\% & 85\% \\
Place & 60\% & 60\% & 60\% & 80\% \\
\bottomrule
\end{tabular*}
\caption{\textbf{Planning and execution ablations.}
Task success rates over 20 trials per task and
configuration. $A_{\mathrm{IK}}$ and $A_{\mathrm{dir}}$
search in the native action space, executing through
inverse kinematics and direct action commands,
respectively. $L_{\mathrm{IK}}$ (ours) and
$L_{\mathrm{dec}}$ optimise latent actions, executing
through geometry-based inverse kinematics and a learned
latent-to-action decoder, respectively.}
\label{tab:planning_ablation}
\end{table}

\subsection{Task Generalisation after Domain Adaptation}\label{sub:dex}

Following DexWM~\cite{goswami2026dexwm}, we evaluate zero-shot task generalisation after task-agnostic fine-tuning on their RoboCasa benchmark, comprising 50 held-out episodes each of Reach, Grasp, and Place with a Franka arm and Allegro hand. To ensure a fair comparison, we restrict both fine-tuning and evaluation to the same single camera view used by DexWM. Our method reconstructs 3D scene geometry from single-view observations and estimates complete goal-hand geometry through kinematic fitting to SAM 3-segmented hand masks \cite{carion2025sam3segmentconcepts}. We fine-tune the model using rank-2 LoRA on task-agnostic exploratory play sequences from kitchens excluded from the evaluation set, without task-specific demonstrations. At deployment, we optimise latent actions using CEM and execute predicted hand motion through kinematic fitting and closed-loop replanning. As shown in Table~\ref{tab:dexterous_comparison}, our method achieves 90\%, 92\%, and 60\% success on Reach, Grasp, and Place, compared with DexWM's reported 72\%, 58\%, and 28\%, respectively. Notably, these improvements are achieved with 20$\times$ fewer candidate evaluations per planning step (256 vs.\ 5,120), using a single H100 GPU compared with DexWM's eight H100s. These results demonstrate dexterous generalisation to unseen tasks following task-agnostic adaptation, with substantially lower planning compute requirements.

\begin{table}
\centering
\footnotesize
\setlength{\tabcolsep}{6pt}
\begin{tabular}{@{}lccc@{}}
\toprule
Method & Reach & Grasp & Place \\
\midrule
DexWM (reported) & 72\% & 58\% & 28\% \\
Ours & 90\% & 92\% & 60\% \\
\bottomrule
\end{tabular}
\caption{\textbf{Dexterous manipulation performance.}
Success rates for Reach, Grasp, and Place. Our method is evaluated over 50 held-out episodes per task, with scene geometry and target poses estimated from images. DexWM results are from the original paper.}
\label{tab:dexterous_comparison}
\end{table}

\begin{figure}
    \centering
    \includegraphics[width=0.9\linewidth]{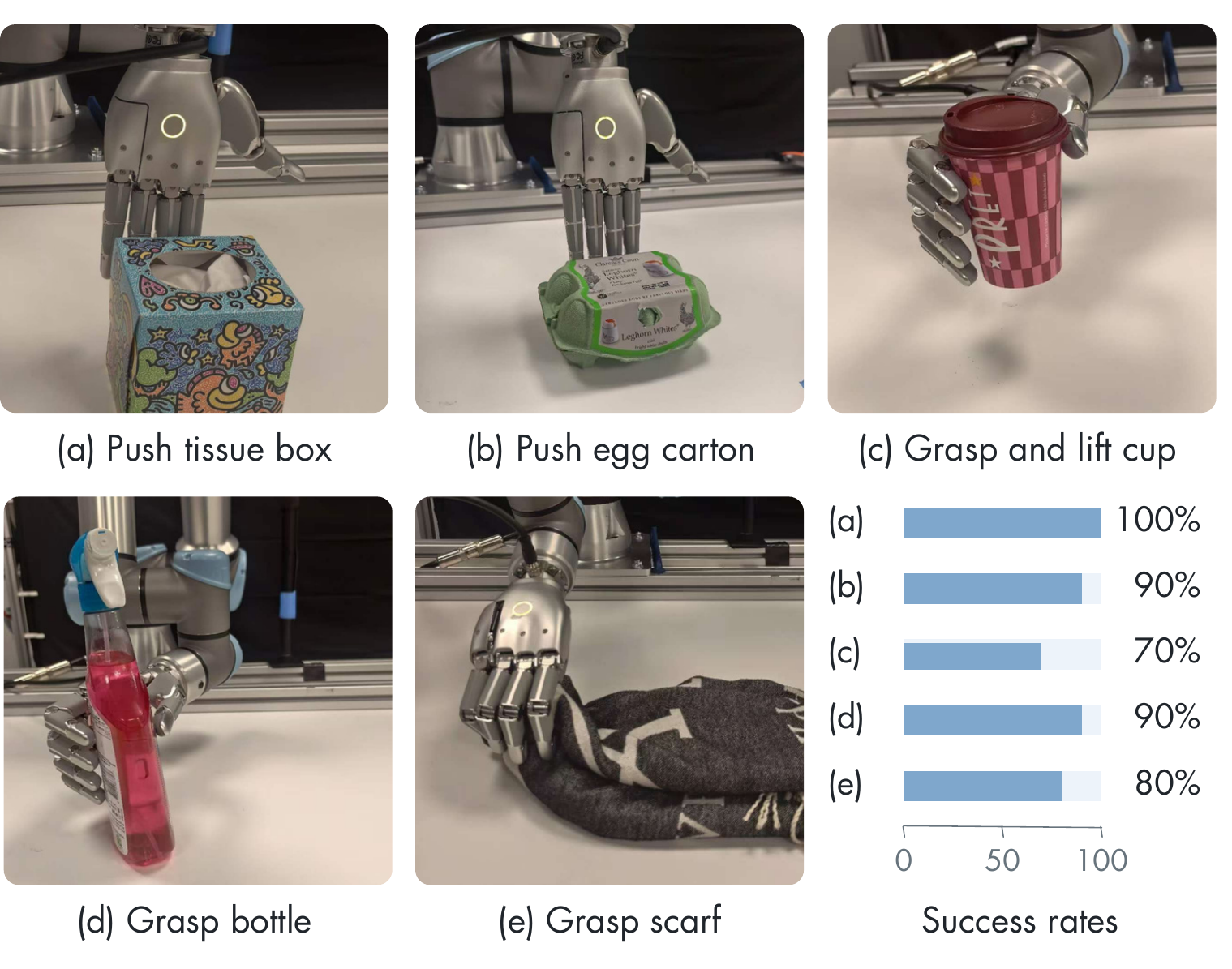}
    \caption{\textbf{Zero-shot real-world dexterous manipulation. }
    (a--e) Representative snapshots of five tasks performed with a Sharpa Wave hand unseen during training, without hand-specific adaptation. Success rates are measured over ten trials per task.}
    \label{fig:real_world}
\end{figure}

\subsection{Real-World Dexterous Manipulation}
\label{sec:real_world}

We deploy the original H2O-pretrained checkpoint on a physical UR7e manipulator equipped with an unseen Sharpa Wave dexterous hand, without further adaptation. Two calibrated RGB-D cameras provide scene point clouds, while hand geometry is reconstructed from measured joint states through forward kinematics. Given a captured goal observation, CEM evaluates candidate latent-action sequences using a Gaussian prior initialised from H2O training statistics. The selected hand prediction is fitted to the Sharpa kinematic model to recover a wrist pose and 22 finger joint targets, which are then executed through closed-loop replanning.

As illustrated in Fig.~\ref{fig:real_world}, we evaluate five tasks with ten trials each: pushing a tissue box, pushing an egg carton, grasping and lifting a cup, grasping a bottle, and grasping a scarf. Success rates are 100\%, 90\%, 70\%, 90\%, and 80\%, respectively, yielding 86\% overall success across 50 trials. These results demonstrate successful zero-shot deployment of the pretrained world model on a physical robot equipped with an unseen dexterous hand.
\section{Limitations \& Conclusion}

Several limitations remain, including the learning of correlations rather than explicit causal interaction mechanisms, contact inconsistencies introduced by retargeting, and limited grasp stability without explicit tactile or force feedback. Addressing these challenges through improved interaction modelling, contact-aware retargeting, and richer physical feedback presents promising directions for future work. Nevertheless, our results demonstrate that a unified latent action space learned from action-unlabelled observations can condition shared predictive dynamics and enable zero-shot control across diverse robot embodiments. DexLatentWorld enables zero-shot manipulation with unseen grippers, outperforms DexWM on dexterous tasks with 20× fewer candidate evaluations, and achieves 86\% success across 50 real-world trials with an unseen dexterous hand, without fine-tuning.

\clearpage
\section*{LLM Usage Statement}
Claude (Anthropic) and ChatGPT (OpenAI) were used to help refine text, code, and figures. All research ideas and scientific contributions are the authors' own. The authors manually reviewed all AI-assisted content and take full responsibility for the work.

{
    \small
    \bibliographystyle{ieeenat_fullname}
    \bibliography{main}
}

\clearpage
\setcounter{page}{1}
\maketitlesupplementary
\appendix
% The CVPR style sets \floatpagefraction and \dblfloatpagefraction to 0.99, so a page of
% tables is only built when it is 99% full; otherwise tables wait for the next \clearpage
% and then come out one per page. The appendix is mostly tables, so allow float pages
% that are 0.3 full. This only affects the pages after this point.
\renewcommand{\floatpagefraction}{0.5}
\renewcommand{\dblfloatpagefraction}{0.5}
% Start float pages at the top instead of centring their tables vertically.
\makeatletter
\setlength{\@fptop}{0pt}\setlength{\@dblfptop}{0pt}
\makeatother

% \section{Rationale}
% \label{sec:rationale}
% %
% Having the supplementary compiled together with the main paper means that:
% %
% \begin{itemize}
% \item The supplementary can back-reference sections of the main paper, for example, we can refer to \cref{sec:intro};
% \item The main paper can forward reference sub-sections within the supplementary explicitly (e.g. referring to a particular experiment);
% \item When submitted to arXiv, the supplementary will already included at the end of the paper.
% \end{itemize}
% %
% To split the supplementary pages from the main paper, you can use \href{https://support.apple.com/en-ca/guide/preview/prvw11793/mac#:~:text=Delete%20a%20page%20from%20a,or%20choose%20Edit%20%3E%20Delete).}{Preview (on macOS)}, \href{https://www.adobe.com/acrobat/how-to/delete-pages-from-pdf.html#:~:text=Choose%20%E2%80%9CTools%E2%80%9D%20%3E%20%E2%80%9COrganize,or%20pages%20from%20the%20file.}{Adobe Acrobat} (on all OSs), as well as \href{https://superuser.com/questions/517986/is-it-possible-to-delete-some-pages-of-a-pdf-document}{command line tools}.
\section*{Appendix}
\label{sec:supp_overview}

This supplementary material lists the settings behind the main paper: data, model and training (\cref{sec:supp_impl}), the evaluation protocols of Sec.~5.1--5.3 with full per-horizon results (\cref{sec:supp_eval}), and planning and control for Sec.~5.4--5.7 (\cref{sec:supp_planning}).

%==========================================================================
\section{Implementation Details}
\label{sec:supp_impl}

\cref{tab:supp_data} summarises the training data and its preprocessing, \cref{tab:supp_model} the architecture, and \cref{tab:supp_train} the training configuration.

\begin{table}[!htb]
\centering
\footnotesize
\setlength{\tabcolsep}{5pt}
\renewcommand{\arraystretch}{1.1}
\begin{tabular}{@{}lp{0.59\linewidth}@{}}
\toprule
\textbf{Operation} & \textbf{Description} \\
\midrule
Source & H2O, split at the sequence level \\
Training split & 76 sequences, 41{,}094 windows \\
Validation split & 16 sequences, 8{,}465 windows \\
Training embodiments & Human hands and 8 retargeted robot hands \\
Robot hands & Ability, Allegro, Inspire, LEAP, OYMotion, Shadow, Wuji, XHand \\
Held-out embodiments & SVH, DG-5F, Dex3 (evaluation only) \\
Retargeting & UniDex-style wrist alignment and fingertip IK \\
Scene entities & Each hand, the object, the table, and 1--5 static distractors \\
Particles & 1{,}024 per entity, area-weighted barycentric sampling \\
Table geometry & RANSAC plane fitted to unprojected depth from 8 frames \\
Window & $K{=}5$ steps, 6 frames apart ($\Delta{=}0.2$\,s) \\
% Centring & Mean of the first frame subtracted \\
Normalisation & Displacements scaled by per-step, per-axis statistics of 4{,}000 training windows \\
Augmentation & None \\
\bottomrule
\end{tabular}
\caption{\textbf{Data and preprocessing.}}
\label{tab:supp_data}
\end{table}
% TODO(Vinay), optional: add a row (or one sentence) on the hand-trajectory refinement of Sec. 4
% (what is optimised, and how the 15.1% / 22.6% reductions are measured).

\begin{table}[!htb]
\centering
\footnotesize
\setlength{\tabcolsep}{5pt}
\renewcommand{\arraystretch}{1.1}
\begin{tabular}{@{}lp{0.66\linewidth}@{}}
\toprule
\textbf{Component} & \textbf{Value} \\
\midrule
\multicolumn{2}{@{}l}{\textbf{Shared PTv3 encoder}} \\
Input & Particle coordinates only \\
Initialisation & From scratch \\
Depths & $(2, 2, 3, 2)$ \\
Channels & $(48, 96, 144, 192)$ \\
Heads & $(2, 4, 6, 8)$ \\
Patch size & 1024 \\
Grid size & 5\,mm \\
Drop path & 0.3 \\
\midrule
\multicolumn{2}{@{}l}{\textbf{Inverse dynamics}} \\
Input tokens & Coarsest-stage tokens of two consecutive observations, a separator, and 4 learned queries \\
Transformer & 2 self-attention layers, width 192, 8 heads \\
Latent action & $z_t \in \mathbb{R}^{4\times 8}$ \\
\midrule
\multicolumn{2}{@{}l}{\textbf{Forward dynamics}} \\
Point features & All encoder stages unpooled to full resolution, concatenated, projected to 192 dimensions \\
Blocks & 1 cross-attention over latent-action tokens, 1 AdaLN self-attention; 8 heads \\
Output & Point-wise head; cumulative displacement at every horizon \\
\midrule
\multicolumn{2}{@{}l}{\textbf{Parameters}} \\
Total & 8.6\,M \\
Encoder & 6.4\,M \\
Inverse dynamics & 1.0\,M \\
Forward dynamics & 1.2\,M \\
\bottomrule
\end{tabular}
\caption{\textbf{Model architecture.} Encoder configurations are ordered by stage.}
\label{tab:supp_model}
\end{table}

\begin{table}[!htb]
\centering
\footnotesize
\setlength{\tabcolsep}{5pt}
\renewcommand{\arraystretch}{1.1}
\begin{tabular}{@{}lp{0.61\linewidth}@{}}
\toprule
\textbf{Setting} & \textbf{Value} \\
\midrule
Optimiser & AdamW \\
Learning rate & $1\times10^{-4}$, constant \\
Weight decay & $10^{-2}$ \\
Gradient clipping & 5 \\
Precision & bfloat16 mixed precision \\
Batch & 4 GPUs $\times$ 4 windows $\times$ 2 embodiments \\
Schedule & 20 epochs (51{,}360 steps) \\
Displacement loss & Huber, $\delta{=}5$ (normalised units), averaged over the 5 horizons \\
Movement weighting & $\kappa{=}1000\,\mathrm{m}^{-1}$, $\tau_m{=}5$\,mm, normalised per scene \\
Contrastive loss & $\lambda_{\mathrm{con}}{=}0.15$, $\tau_c{=}0.1$ \\
Positives & First-transition latents of one window under two embodiments \\
Negatives & Other windows in the per-GPU batch; at most one window per sequence \\
Checkpoint & Lowest validation Mover EPE, evaluated every epoch \\
Training GPUs & 4 GH200 \\
Training time & $\approx$16 hours \\
\bottomrule
\end{tabular}
\caption{\textbf{Training configuration.}}
\label{tab:supp_train}
\end{table}

%==========================================================================
\section{Evaluation Details}
\label{sec:supp_eval}

\begin{figure*}[t]
    \centering
    \includegraphics[width=0.9\textwidth]{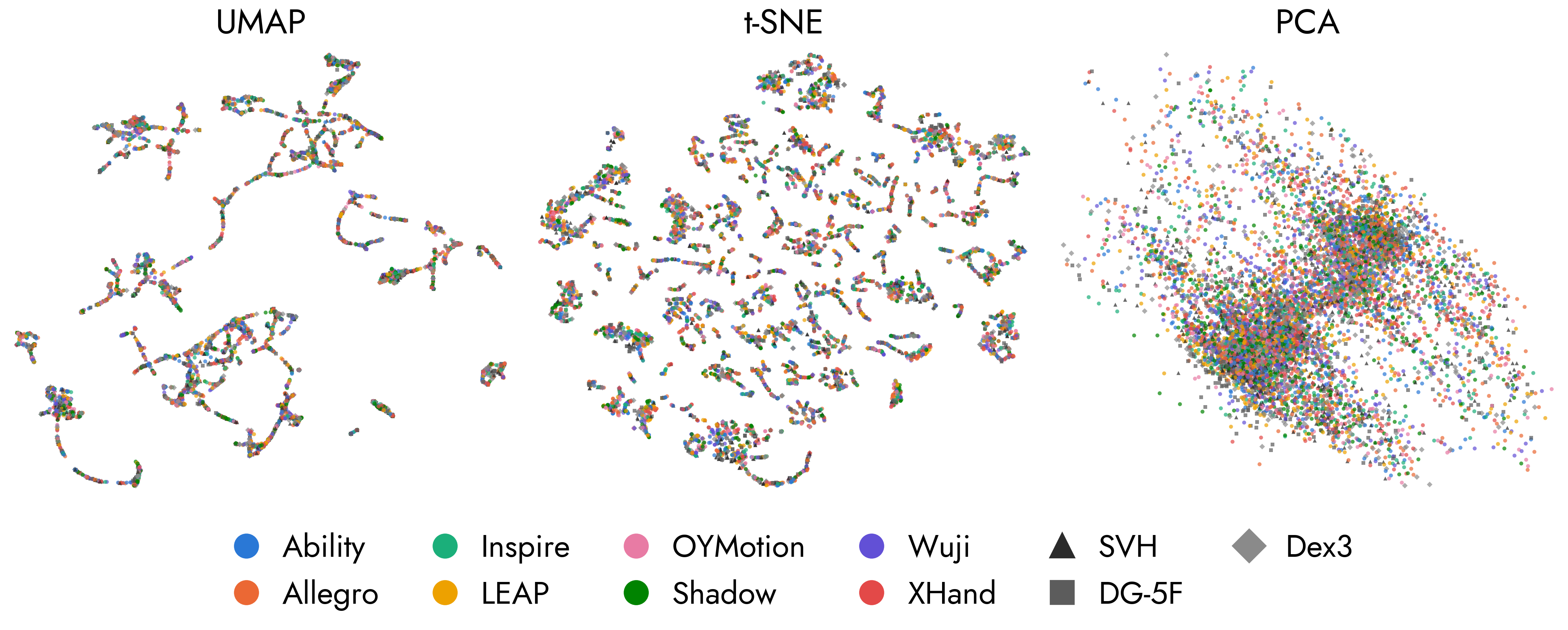}
    \caption{\textbf{Latent action embeddings across embodiments.} UMAP, t-SNE, and PCA visualisations (left to right) of 8{,}465 first-step latent actions (32 dimensions, comprising four 8-dimensional tokens) inferred from held-out H2O validation sequences. Each window is encoded under one randomly assigned embodiment, with colours indicating the eight training robot hands and grey markers the three held-out hands (SVH, DG-5F, Dex3). The embeddings overlap substantially across embodiments. Local structure largely reflects temporal continuity, with consecutive windows from the same manipulation appearing nearby across different embodiments.}
    \label{fig:supp_embeddings}
\end{figure*}

{\interlinepenalty=10000 % keep this short paragraph in one piece
\cref{tab:supp_eval} gives the protocols for Sec.~5.1--5.3.
\cref{tab:supp_latent_full} extends Tab.~1 of the main paper to all five horizons: correct conditioning gives the lowest hand and object error at every step. \cref{fig:supp_embeddings} complements the embodiment probe: latent actions of different embodiments overlap, with no clear morphology-specific clusters.\par}

\begin{table}[!htb]
\centering
\footnotesize
\setlength{\tabcolsep}{5pt}
\renewcommand{\arraystretch}{1.1}
\begin{tabular}{@{}lp{0.66\linewidth}@{}}
\toprule
\textbf{Setting} & \textbf{Description} \\
\midrule
\multicolumn{2}{@{}l}{\textbf{Latent conditioning (Sec.~5.1)}} \\
Windows & 4{,}000 validation windows (250 per sequence), each evaluated under one of the nine training embodiments \\
Zero & All latents set to zero \\
Shuffled & Each window receives the latents of a different window in its batch (8 windows, all from different sequences) \\
Fixed & Latents of a single donor window applied to every window \\
\midrule
\multicolumn{2}{@{}l}{\textbf{Cross-decoding (Sec.~5.3)}} \\
Windows & 256 validation windows (16 per sequence), valid for all eleven robot embodiments \\
Metric & Mover EPE averaged over the 5 horizons \\
No-motion baseline & 49.5\,mm: zero displacement, pooled over the same windows and embodiments \\
\midrule
\multicolumn{2}{@{}l}{\textbf{Embodiment probe (Sec.~5.3)}} \\
Classifier & Standardised, $\ell_2$-regularised logistic regression \\
Data & 822 windows $\times$ 11 robot embodiments (9{,}042 samples) \\
Split & 6{,}820 for training; 2{,}222 for testing, from 4 held-out validation sequences \\
Inputs compared & Latent actions of all 5 steps vs.\ mean-pooled coarsest-stage PTv3 features of the first frame \\
\bottomrule
\end{tabular}
\caption{\textbf{Evaluation protocols} for Sec.~5.1--5.3.}
\label{tab:supp_eval}
\end{table}

\begin{table*}[!t]
\centering
\footnotesize
\setlength{\tabcolsep}{2pt}
\resizebox{\textwidth}{!}{%
\begin{tabular}{@{}l*{18}{c}@{}}
\toprule
& \multicolumn{3}{c}{Step 1}
& \multicolumn{3}{c}{Step 2}
& \multicolumn{3}{c}{Step 3}
& \multicolumn{3}{c}{Step 4}
& \multicolumn{3}{c}{Step 5}
& \multicolumn{3}{c}{Mean} \\
\cmidrule(lr){2-4}
\cmidrule(lr){5-7}
\cmidrule(lr){8-10}
\cmidrule(lr){11-13}
\cmidrule(lr){14-16}
\cmidrule(lr){17-19}
Prediction
& Hand & Object & Scene
& Hand & Object & Scene
& Hand & Object & Scene
& Hand & Object & Scene
& Hand & Object & Scene
& Hand & Object & Scene \\
\midrule
No motion (baseline)
& 25.97 & 23.18 & 0.00
& 41.80 & 35.30 & 0.00
& 55.16 & 44.84 & 0.00
& 66.66 & 53.66 & 0.00
& 76.78 & 61.50 & 0.00
& 53.28 & 43.70 & 0.00 \\
\midrule
Correct latents
& \textbf{4.61} & \textbf{8.51} & 0.39
& \textbf{6.54} & \textbf{11.49} & 0.67
& \textbf{8.12} & \textbf{13.80} & 0.83
& \textbf{9.55} & \textbf{16.02} & 1.10
& \textbf{10.87} & \textbf{18.12} & 1.34
& \textbf{7.94} & \textbf{13.59} & 0.87 \\
Zero latents
& 25.94 & 23.03 & 0.28
& 41.74 & 35.11 & 0.57
& 55.06 & 44.57 & 0.73
& 66.47 & 53.25 & 0.97
& 76.66 & 61.18 & 0.84
& 53.17 & 43.43 & 0.68 \\
Shuffled latents
& 34.90 & 26.49 & 0.39
& 59.11 & 42.32 & 0.66
& 78.74 & 54.40 & 0.80
& 94.58 & 64.29 & 1.03
& 107.72 & 72.29 & 1.21
& 75.01 & 51.96 & 0.82 \\
Fixed latents
& 30.34 & 24.33 & 0.34
& 49.21 & 38.68 & 0.60
& 63.47 & 48.98 & 0.70
& 74.37 & 57.11 & 0.92
& 84.06 & 64.81 & 0.94
& 60.29 & 46.78 & 0.70 \\
\bottomrule
\end{tabular}%
}
\caption{\textbf{Latent conditioning at every prediction horizon.}
Full version of Tab.~1 of the main paper: particle EPE in mm ($\downarrow$) at each of the five rollout steps on our H2O validation split (nine training embodiments).
Hand and object errors are evaluated on points with ground-truth displacement $>5$\,mm, while scene errors include static table and distractor points.
No motion assumes zero displacement. Other rows use correct, zeroed, shuffled, or fixed-donor latent actions.}
\label{tab:supp_latent_full}
\end{table*}

%==========================================================================
\section{Planning and Control}
\label{sec:supp_planning}

{\interlinepenalty=10000 % keep this short paragraph in one piece
All experiments plan with CEM over a five-step, 160-dimensional latent-action sequence, initialised from a diagonal Gaussian fitted to encoder latents: those of H2O training windows for the pretrained model, and those of the adaptation sequences for fine-tuned models. CEM starts from the prior mean and standard deviation, with an element-wise floor of 0.05. Planning is receding-horizon.
\cref{tab:planning_config} lists the settings that differ between the gripper and dexterous-hand experiments.\par}

\begin{table*}[!t]
\centering
\footnotesize
\renewcommand{\arraystretch}{1.1}
\begin{tabular}{@{}lp{0.37\textwidth}p{0.37\textwidth}@{}}
\toprule
\textbf{Parameter} & \textbf{Gripper (Sec.~5.4)} & \textbf{Dexterous hand (Sec.~5.5--5.6)} \\
\midrule
CEM budget
& 64 samples $\times$ 4 iterations; 128 $\times$ 6 in the push phase
& 64 samples $\times$ 4 iterations for all tasks \\
Elite fraction
& 10\%
& 20\% \\
Search initialisation
& Warm-started from the previous solution
& Restarted from the fitted Gaussian prior \\
Selected solution
& Final elite mean
& Lowest-cost candidate across iterations \\
% Hold candidate
% & Plan must improve on it by 0.5\,mm
% & Executed if it has the lowest cost \\
Executed horizon
& Task-dependent % TODO(Vinay): say which predicted step each task executes, or drop this row
& First predicted step \\
Motion recovery
& Gripper centroid; rigid alignment for reaching
& Palm alignment and finger-wise IK \\
Control
& Task-specific Cartesian targets
& Wrist targets and finger PD control \\
\bottomrule
\end{tabular}
\caption{\textbf{Planning and control configurations.} Both settings use a five-step planning horizon and a 160-dimensional latent-action sequence.}
\label{tab:planning_config}
\end{table*}

\subsection{Zero-Shot Gripper Manipulation (Sec.~5.4)}
\label{sec:supp_gripper}

\cref{tab:supp_gripper_setup} gives the shared setup and \cref{tab:supp_gripper_tasks} the per-task goals, replanning budgets and success criteria.

\begin{table}[!htb]
\centering
\footnotesize
\setlength{\tabcolsep}{5pt}
\renewcommand{\arraystretch}{1.1}
\begin{tabular}{@{}lp{0.66\linewidth}@{}}
\toprule
\textbf{Setting} & \textbf{Value} \\
\midrule
Robot & Franka Panda in RoboSuite \\
Grippers & Robotiq140 (Cube Stacking, Milk Transport); WipingGripper (Push Cube, Reach + Push Puck) \\
Control & Operational-space position control, 20\,Hz \\
Scene particles & Sampled from simulator geometry (gripper, object, table or bins); identities kept across replans \\
Goal & Stacking, transport: object particles placed at the target (transport via lift, carry and lower waypoints); push: pre-push approach pose, then push towards a target 200\,mm ahead \\
Planner & CEM, 64 samples $\times$ 4 iterations (128 $\times$ 6 in the push phase), 10\% elites \\
Executed command & Mean predicted gripper displacement, clipped to 60\,mm, tracked over 25 control steps \\
\midrule
\multicolumn{2}{@{}l}{\textbf{PointWorld baseline}} \\
Model & Released checkpoint \\
Observation & Same scenes through two RGB-D cameras \\
Candidates & Cubic B-spline gripper trajectories, 4 knots \\
Scoring & Predicted object and scene motion under our cost, CEM budget, executor and episode limits \\
\bottomrule
\end{tabular}
\caption{\textbf{Zero-shot gripper setup.}}
\label{tab:supp_gripper_setup}
\end{table}

\begin{table*}[!t]
\centering
\footnotesize
\setlength{\tabcolsep}{4pt}
\renewcommand{\arraystretch}{1.1}
\begin{tabular}{@{}ll>{\raggedright\arraybackslash}p{0.30\textwidth}>{\raggedright\arraybackslash}p{0.12\textwidth}>{\raggedright\arraybackslash}p{0.21\textwidth}@{}}
\toprule
\textbf{Task} & \textbf{Gripper} & \textbf{Goal and cost} & \textbf{Replans} & \textbf{Success criterion} \\
\midrule
Cube Stacking & Robotiq140
& A scripted grasp first lifts the cube. Distance of the predicted object particles to the goal at the final horizon.
& 20
& Lifted, within 50\,mm of the lower cube in the plane, and in contact \\
Milk Transport & Robotiq140
& A scripted grasp first lifts the carton. Same cost over three goals: lift above the bin walls, carry above the target bin, lower into it.
& 5 (lift)\newline 20 (carry)\newline 20 (lower)
& Carton within 20\,mm of its target position \\
Push Cube & WipingGripper
& The gripper starts behind the cube. Negated ManiSkill PushCube dense reward on the predicted motion.
& 150
& Cube within 40\,mm of a target 200\,mm ahead \\
Reach + Push Puck & WipingGripper
& Approach to a pre-push pose with a table-penetration barrier, then the same push cost.
& 50 (approach)\newline 15 (push)
& Puck within 40\,mm of a target 200\,mm ahead \\
\bottomrule
\end{tabular}
\caption{\textbf{Gripper tasks.} Replans are the maximum per phase.}
\label{tab:supp_gripper_tasks}
\end{table*}

\subsection{Dexterous Manipulation (Sec.~5.5--5.6)}
\label{sec:supp_dex}

{\interlinepenalty=10000 % keep this short paragraph in one piece
Both settings use a Franka arm and Allegro hand at 10\,Hz under operational-space wrist control and joint PD finger control.
\cref{tab:supp_adapt} describes the task-agnostic adaptation, \cref{tab:supp_dex} the differences between Sec.~5.5 and Sec.~5.6, and \cref{tab:supp_dex_plan} the planner, execution and evaluation.\par}

\begin{table}[!t]
\centering
\footnotesize
\setlength{\tabcolsep}{5pt}
\renewcommand{\arraystretch}{1.1}
\begin{tabular}{@{}lp{0.68\linewidth}@{}}
\toprule
\textbf{Setting} & \textbf{Value} \\
\midrule
\multicolumn{2}{@{}l}{\textbf{Play data}} \\
Clips & 12 eight-second clips in each of 13 non-evaluation kitchens \\
Wrist motion & Random paths \\
Finger motion & Random, with grasp-like closures in half of the motions; never directed at the object \\
Split & 84 clips (7 kitchens) for gradients; 72 clips (6 kitchens) for early stopping \\
\midrule
\multicolumn{2}{@{}l}{\textbf{Observations}} \\
Scene points & RGB-D back-projected, cropped, voxelised at 1.5\,cm; 5{,}120 points \\
Robot removal & Points within 1.5\,cm of the robot surface (forward kinematics) \\
Hand particles & 1{,}024, placed by forward kinematics \\
\midrule
\multicolumn{2}{@{}l}{\textbf{Correspondences}} \\
Tracker & CoTracker3, 1{,}024 scene points per window \\
Filtering & Lifted to 3D with depth; kept if visible at all 5 horizons \\
Supervision & Hand particles always; untracked scene points are input only \\
\midrule
\multicolumn{2}{@{}l}{\textbf{Fine-tuning}} \\
Adapters & LoRA, rank 2, $\alpha{=}2$, on every linear layer of the inverse and forward dynamics models; encoder frozen \\
Uncertainty & Linear head, per-particle log-uncertainty \\
Optimiser & AdamW, learning rate $5\times10^{-4}$, cosine schedule \\
Batch & 4 windows per update \\
Updates & At most 3{,}000 \\
Early stopping & Held-out hand error every 100 updates; stop after 3 checks without a 0.5\,mm improvement \\
\bottomrule
\end{tabular}
\caption{\textbf{Task-agnostic domain adaptation.}}
\label{tab:supp_adapt}
\end{table}

\begin{table}[!t]
\centering
\footnotesize
\setlength{\tabcolsep}{5pt}
\renewcommand{\arraystretch}{1.1}
\begin{tabular}{@{}l>{\raggedright\arraybackslash}p{0.27\linewidth}>{\raggedright\arraybackslash}p{0.31\linewidth}@{}}
\toprule
\textbf{Setting} & \textbf{Sec.~5.5} & \textbf{Sec.~5.6} \\
\midrule
Purpose & Ablation & DexWM benchmark \\
Cameras & Two RGB-D views & DexWM's single view \\
Goal & Target hand state & Hand fitted to the goal image \\
Episodes & 20 per task & All 50 per task \\
Configurations & $A_{IK}$, $A_{dir}$, $L_{IK}$, $L_{dec}$ & $L_{IK}$ \\
Held-out hand error & 60.6 $\rightarrow$ 17.8\,mm & 65.3 $\rightarrow$ 22.8\,mm \\
\bottomrule
\end{tabular}
\caption{\textbf{Differences between the two dexterous settings.} Held-out hand error is before $\rightarrow$ after fine-tuning. The 20 episodes of Sec.~5.5 are the first evaluation demonstrations not already solved at the start or by holding still. Sec.~5.6 runs on a single H100 GPU.}
\label{tab:supp_dex}
\end{table}

% \begin{figure}
%     \centering
%     \includegraphics[width=1\linewidth]{Images/scaling_figure.png}
%
%     \caption{\textbf{Data efficiency of world-model adaptation to a new environment distribution.}
%     We fine-tune rank-2 LoRA adapters on the pretrained world model using $N\in\{10,20,40,80,160\}$ task-agnostic play clips. Performance is measured by mover endpoint error (EPE) at the 1 s prediction horizon, averaged over points with ground-truth displacement exceeding 5 mm. We train on data from eight kitchens, validate on two separate kitchen environments, and evaluate on 60 clips from three unseen test kitchens. Hand-point ground truth is obtained from forward kinematics and scene-point motion from CoTracker3; 98\% of moving points belong to the hand. Shaded regions indicate 95\% bootstrap confidence intervals over test clips. Error decreases substantially up to 40 clips, with diminishing improvements thereafter.}
%
%     \label{fig:supp_scaling}
% \end{figure}

\begin{table*}[!t]
\centering
\footnotesize
\renewcommand{\arraystretch}{1.1}
\begin{tabular}{@{}lp{0.78\textwidth}@{}}
\toprule
\textbf{Component} & \textbf{Description} \\
\midrule
\multicolumn{2}{@{}l}{\textbf{Planning}} \\
CEM & $64\times4=256$ candidate sequences per planning step for every task; 20\% elites \\
Horizon weights & $(0.05, 0.10, 0.15, 0.20, 0.50)$ \\
\midrule
\multicolumn{2}{@{}l}{\textbf{Task costs}} \\
Reach & Distance of the predicted hand particles to the goal hand, finger-surface points weighted twice relative to palm points; penalty on predicted scene motion \\
Grasp & Same hand distance; first targets a pose 200\,mm above the goal hand until the hand is within 50\,mm of it \\
Place & One-sided Chamfer distance from the predicted object particles to the goal object, a hand--object attachment term, and a down-weighted hand term \\
\midrule
\multicolumn{2}{@{}l}{\textbf{Execution}} \\
Executed horizon & First predicted step \\
Palm pose & Kabsch alignment of the predicted palm particles \\
Finger joints & Fitted per finger by bounded nonlinear least squares within joint limits \\
Limits & Each target held for 0.5\,s; at most 60\,mm of wrist translation per replan; up to 80 replans \\
\midrule
\multicolumn{2}{@{}l}{\textbf{Action maps (Sec.~5.5)}} \\
Networks & Two-layer MLPs (hidden width 512, LayerNorm, SiLU) between a 22-dimensional hand action and one latent step \\
Hand action & Wrist translation, wrist rotation vector, 16 finger joint increments \\
Training data & Achieved hand motion of the play data \\
Action-space search & Increments sampled from a per-step Gaussian fitted to the play actions \\
\midrule
\multicolumn{2}{@{}l}{\textbf{Goal hands from images (Sec.~5.6)}} \\
Segmentation & SAM~3 with three generic prompts (``black-coloured robot hand'', ``robot'', ``robotic hand'') \\
Kinematic fit & Arm and hand joint angles fitted to the masked depth points with a batched, differentiable silhouette and visible-depth objective \\
Initialisations & 512, of which the best 32 are refined; ground-truth goal joint angles are never used \\
\midrule
\multicolumn{2}{@{}l}{\textbf{Success criteria} (DexWM's definitions, held for 10 consecutive control steps)} \\
Reach & Hand within 0.15\,m of the target \\
Grasp & Hand within 0.20\,m of the object and in contact \\
Place & Object within 0.10\,m of its target \\
\bottomrule
\end{tabular}
\caption{\textbf{Dexterous planning, execution and evaluation.}}
\label{tab:supp_dex_plan}
\end{table*}

% TODO(Vinay), recommended: 
\subsection{Real-World Dexterous Manipulation (Sec.~5.7)}
\label{sec:supp-real-world}
% The supplementary currently says nothing about Sec. 5.7. A table in the same style would cover:
% CEM budget and elite fraction, control rate, how the goal observation is captured, how the two
% RGB-D views are fused and cropped, and the success criterion for each of the five tasks.

\paragraph{Hardware and observations.}
We deploy our H2O-pretrained model on a UR7e arm with a Sharpa Wave hand (22 finger joints), using a single NVIDIA RTX 4090 GPU (24~GB) for inference. Two table-mounted Intel RealSense D435 cameras provide first- and third-person views, respectively. Depth observations from both calibrated cameras are back-projected into the robot base frame, concatenated, and cropped to an axis-aligned workspace bounding box. After voxel filtering and removal of points near the FK robot surface, the scene is sampled to 5,120 points. We add 1,024 fixed hand-surface samples placed by FK from measured joints, preserving hand-point correspondences across observations. A fixed rotation maps observations into the training coordinate convention and predicted displacements back into the robot base frame.

\paragraph{Goal capture and planning.}
Goals are captured by manually arranging the robot and recording calibrated depth and joint states. Target hand points are reconstructed by FK in the robot base frame, requiring consistent object placement.
CEM searches five-step latent-action sequences from the H2O-derived Gaussian prior, selecting the lowest-cost sampled trajectory across iterations or a zero-displacement alternative.
The cost measures horizon-weighted distances to corresponding goal-hand points, weighting finger-surface points twice relative to palm points.

\paragraph{Execution and evaluation.}
Predicted hand points are fitted to the Sharpa kinematics to recover wrist and 22 finger targets, executed through Cartesian arm motions and interpolated finger commands. We execute the first three targets sequentially using UR Cartesian linear motions and minimum-jerk finger interpolation through the Sharpa SDK. Each target is checked against fresh joint feedback, controller IK, and motion limits.
The system alternates observation, planning, motion, and settling, rather than running at a fixed planning rate.
Table~\ref{tab:real-world-deployment} lists the key settings and physical success criteria. Each task is evaluated over ten trials.

\begin{table}[!htb]
\centering
\footnotesize
\setlength{\tabcolsep}{3pt}
\renewcommand{\arraystretch}{1.08}
\begin{tabular}{@{}>{\raggedright\arraybackslash}p{0.32\columnwidth}p{\dimexpr0.68\columnwidth-2\tabcolsep\relax}@{}}
\toprule
\textbf{Setting} & \textbf{Value} \\
\midrule
Depth & $848\times480$, 30~Hz; emitters enabled; range 0.2--2.0~m \\
Filtering & 15~mm voxels; 15~mm robot-removal margin \\
Model input & 1,024 hand points + 5,120 scene points \\
CEM & 64 samples $\times$ 4 iterations; 20\% elites \\
Planning horizon & 5 steps; $4\times8$ latent dimensions per step (160 in total) \\
Executed horizon & The first 3 predicted targets \\
Control & 100~Hz finger commands; replanning after motion and settling \\
Translation limits & 60~mm per target; 0.02~m/s \\
\midrule
\multicolumn{2}{@{}l}{\textbf{Task success criteria}} \\
Push tissue box;\newline Push egg carton
& Object reaches the designated tabletop target region \\
Grasp and lift cup
& Cup is grasped, lifted $\geq5$~cm clear of the table
  and held for 3~s without dropping \\
Grasp bottle;\newline Grasp scarf
& Object successfully grasped \\
\bottomrule
\end{tabular}
\caption{\textbf{Real-world deployment and evaluation.} Each of the five tasks is evaluated over ten trials.}
\label{tab:real-world-deployment}
\end{table}


\end{document}